# Sequential Update of Bayesian Network Structure

**Nir Friedman**
University of California
Computer Science Division
Berkeley, CA 94720
nir@cs.berkeley.edu

**Moises Goldszmidt**
SRI International
333 Ravenswood Ave, EK329
Menlo Park, CA 94025
moises@erg.sri.com

## Abstract

There is an obvious need for improving the performance and accuracy of a Bayesian network as new data is observed. Because of errors in model construction and changes in the dynamics of the domains, we cannot afford to ignore the information in new data. While sequential update of parameters for a fixed structure can be accomplished using standard techniques, sequential update of network structure is still an open problem. In this paper, we investigate sequential update of Bayesian networks were both parameters and structure are expected to change. We introduce a new approach that allows for the flexible manipulation of the tradeoff between the quality of the learned networks and the amount of information that is maintained about past observations. We formally describe our approach including the necessary modifications to the scoring functions for learning Bayesian networks, evaluate its effectiveness through and empirical study, and extend it to the case of missing data.

## 1 Introduction

Recently, there has been a great deal of effort in developing methods for learning Bayesian networks from data for density estimation, data analysis, and pattern classification (see [7] for a tutorial and an overview). This body of work, which includes both theoretical and experimental results, has concentrated mostly on *batch* learning methods. In this setting, the total corpus of data is fully available to the learning algorithm which outputs a model after multiple inspections of the data.

In this paper we study the problem of *sequential* update of Bayesian networks. This problem is different from batch learning in two aspects: (1) the learning procedure receives the data as a "read-once" stream of observations, and (2) the learning procedure has to output a model, based on the observations seen so far, at various time points (possibly after each observation is made).

Sequential update is a crucial capability for building adaptable systems that can overcome errors in their initial model, and that can adapt to changes in the underlying distribution. We are especially interested in sequential update in situations where the learning procedure cannot store all of the past observations (or a complete summary of it) in main memory. Examples of such situations include monitoring systems which collect data over extended periods of time or embedded systems with limited memory capabilities. Memory constraints also arise in data mining applications that usually involve massive amount of data that must be kept on secondary storage. In these applications repeated inspection of the data is infeasible.

We claim that effective sequential update of structure involves a tradeoff between the quality of the learned networks and the amount of information that is maintained about past observations. We consider three approaches to sequential update. Two of these approaches lie on the extremes of the spectrum. The third approach, which is new, allows for a flexible manipulation of the tradeoff.

The *naive* approach stores all the previously seen data, and repeatedly invokes a batch learning procedure after each new datum is recorded. This approach can use all of the information provided so far, and thus is essentially optimal in terms of the quality of the networks it can induce. This approach, however, requires vast amount of memory to store the entire corpus of data.[1]

We can attempt to avoid the overhead of storing all of the previously seen data instances by summarizing them using the model we have learned so far. This approach essentially assumes that the set of data instances being summarized are distributed according to the probability measure described by the current model. This approach, which we call *MAP* (for reasons that will become clear in Section 3.1), is space-efficient. Unfortunately, by using the current model as a summary of past data, we strongly bias our learning procedure towards that model. As a result, after some number of iterations, this approach locks itself into a particular model and stops adapting to new data.

---

[1] Of course, it suffices to keep counts of the number of times each distinct case was observed so far. This, however, is also impractical, since the number of distinct cases is exponential in the number of variables of interest.

The third approach, which we call *incremental*, provides a middle ground between the extremes defined by the naive and the MAP approaches. Moreover, it allows flexible choices in the tradeoff between space and quality of the induced networks. The incremental approach interleaves steps in a search process, to find "good" models, with the incorporation of new data. This approach focuses its resources on keeping track of just enough information to make the next decision in the search process. The basic strategy is to maintain a set of network candidates that are on the *frontier* of this process. This set contains all of the networks that are deemed most promising at the current time. The procedure also keeps track of the required information needed for evaluating candidates on the frontier. As we shall see, this information can be maintained in a space-efficient manner. As each new data case arrives, the procedure updates the information stored in memory, and invokes the search process to check whether one of the networks in the frontier is deemed more suitable than the current model. (We always assume that the current network is on the frontier.) If this is the case, it adopts the new model as current model, and updates the search frontier. As we shall see, the amount of space required by the procedure is related to the size of the search frontier.

The dynamics in the recording of information about the data raises a fundamental question with respect to the scoring functions commonly used to evaluate different models. All of the scoring functions proposed in the literature assume that alternative candidates are evaluated with respect to the same training data. In the incremental learning procedure, however, we can start recording the information for different candidates at different times. We propose modifications to the MDL score and the Bayesian score that deal with this complication. We empirically evaluate these "extended" scoring functions in conjunction with the incremental learning procedure.

Finally, we examine how to extend these methods to deal with incomplete data in sequential update. To this end, we propose a combination of two generalizations of the *expectation maximization* algorithm: *incremental EM* [12] and *model-selection EM* [4].

The rest of this paper is organized as follows: In Section 2, we briefly review the current practice of learning Bayesian networks. In Section 3, we describe our approach for incremental update and develop the necessary theoretical foundations. In Section 4, we perform an empirical evaluation of the three methods described above, In Section 5, we introduce the extension to missing data. We conclude in Section 6 with a discussion of related work, a summary of the main the results of this paper, and our plans for future work.

## 2 Learning Bayesian Networks: The Batch Method

Consider a finite set $\mathbf{U} = \{X_1, \ldots, X_n\}$ of discrete random variables where each variable $X_i$ may take on values from a finite set, denoted by $Val(X_i)$. We use capital letters, such as $X, Y, Z$, for variable names and lowercase letters $x, y, z$ to denote specific values taken by those variables. Sets of variables are denoted by boldface capital letters $\mathbf{X}, \mathbf{Y}, \mathbf{Z}$, and assignments of values to the variables in these sets are denoted by boldface lowercase letters $\mathbf{x}, \mathbf{y}, \mathbf{z}$ (we use $Val(\mathbf{X})$ in the obvious way). Finally, let $P$ be a joint probability distribution over the variables in $\mathbf{U}$, and let $\mathbf{X}, \mathbf{Y}, \mathbf{Z}$ be subsets of $\mathbf{U}$. The sets $\mathbf{X}$ and $\mathbf{Y}$ are *conditionally independent* given $\mathbf{Z}$ if for all $\mathbf{x} \in Val(\mathbf{X}), \mathbf{y} \in Val(\mathbf{Y}), \mathbf{z} \in Val(\mathbf{Z})$, $P(\mathbf{x} \mid \mathbf{z}, \mathbf{y}) = P(\mathbf{x} \mid \mathbf{z})$ whenever $P(\mathbf{y}, \mathbf{z}) > 0$.

A *Bayesian network* is an annotated directed acyclic graph that encodes a joint probability distribution of a set of random variables $\mathbf{U}$. Formally, a Bayesian network for $\mathbf{U}$ is a pair $B = \langle G, \Theta \rangle$. The first component, namely $G$, is a directed acyclic graph whose vertices correspond to the random variables $X_1, \ldots, X_n$, and whose edges represent direct dependencies between the variables. The graph $G$ encodes the following set of independence assumptions: each variable $X_i$ is independent of its non-descendants given its parents in $G$. The second component of the pair, namely $\Theta$, represents the set of parameters that quantifies the network. It contains a parameter $\theta_{x_i|\mathbf{pa}(x_i)} = P_B(x_i \mid \mathbf{pa}(x_i))$ for each possible value $x_i$ of $X_i$, and $\mathbf{pa}(x_i)$ of $\mathbf{pa}(X_i)$, where $\mathbf{pa}(X_i)$ denotes the set of parents of $X_i$ in $G$. A Bayesian network $B$ defines a unique joint probability distribution over $\mathbf{U}$ given by: $P_B(X_1, \ldots, X_n) = \prod_{i=1}^n P_B(X_i \mid \mathbf{pa}(X_i))$.

The problem of learning a Bayesian network can be stated as follows. Given a *training set* $D = \{\mathbf{u}_1, \ldots, \mathbf{u}_N\}$ of instances of $\mathbf{U}$, find a network $B$ that *best matches* $D$. The common approach to this problem is to introduce a scoring function (or a *score*) that evaluates the "fitness" of networks with respect to the training data, and then to search for the best network (according to this score). The two main scoring functions commonly used to learn Bayesian networks are the *Bayesian* score [3, 8], and the one based on the principle of *minimal description length* (MDL) [9, 5]. These scores are asymptotically equivalent as the sample size increases. Furthermore they are both asymptotically correct: with probability equal to one the learned distribution converges to the underlying distribution as the number of samples increases [7, 6].

In this paper we use the MDL score described in [5], which we denote as $S_{MDL}$, and the "BDe" variant of the Bayesian introduced by Heckerman et. al [8], which we denote as $S_{BDe}$. Details about these scores for batch learning can be found in [5, 8]. What is of interest for the purposes of this paper is to understand what information (from the training data) is needed to compute these scores.

When the data is *complete*, namely, each instance assigns values to all the variables of interest, both scores have two attractive properties. The first property is that for any fixed network structure $G$, there is a closed-form formula for finding the optimal parameters that maximize the score. Moreover, these parameters can be extracted from *sufficient statistics* for the structure $G$. To understand the notion of sufficient statistics it is convenient to introduce additional notation. Let $N_\mathbf{X}^D(\mathbf{x})$ be the number of instances in $D$ where $\mathbf{X} = \mathbf{x}$. Let $\hat{N}_\mathbf{X}^D$ be the vector of numbers $N_\mathbf{X}^D(\mathbf{x})$ for all values of $\mathbf{X}$ (from now on, we omit the superscript and the subscript of $N_\mathbf{X}^D$ whenever they are clear from the context.) We call the vector $\hat{N}_\mathbf{X}$ the *sufficient statistics* of $\mathbf{X}$. As it turns out, we the optimal choice of parameters $\theta_{X_i|\mathbf{pa}(X_i)}$ is a function of $\hat{N}_{X_i,\mathbf{pa}(X_i)}$; see, for example, [7]. Since the selection of parameters has a closed form, we can focus on choosing the best structure $G$ for the network. The parameters are then easily computed from these sufficient statistics.

The second property is *decomposability* of the score assigned to a structure. This means that the score $S(G \mid D)$ assigned to a structure $G$, in the context of a dataset $D$, has the general form:

$$S(G \mid D) = \sum_i S(X_i, \mathbf{pa}(X_i)) \tag{1}$$

where $S(X_i, \mathbf{pa}(X_i))$ is the *local* score evaluating how good is the choice of parents for $X_i$ (as determined by $G$). Let a *family* be the set composed by $X_i$ and its parents. The local score $S(X_i, \mathbf{pa}(X_i))$ depends only on the sufficient statistics for this family: $\hat{N}_{X_i,\mathbf{pa}(X_i)}$. Thus, if $D$ and $D'$ are such that $\hat{N}_{X_i,\mathbf{pa}(X_i)}^D = \hat{N}_{X_i,\mathbf{pa}(X_i)}^{D'}$, then the local score of this particular family will be the same. The direct benefit of this property is computational. To evaluate the effect in the score that the addition (or removal) of an arc in $G$ will have, we need only to recompute the local score of the families affected.

In conclusion, for the purposes of learning using either the MDL or the Bayesian score, all the required information about the training data $D$ is summarized by a set of sufficient statistics. These statistics are of the form $\hat{N}_{X_i,\mathbf{pa}(X_i)}$ for choices of $\mathbf{pa}(X_i)$ in the set of networks considered during the search.

## 3  Sequential Update of Bayesian Networks

Sequential update of Bayesian networks is an *on-line* learning problem. At each iteration $n$, procedure receives a new data instance $\mathbf{u}_n$, and then produces the next hypothesis $B_{n+1}$. This estimate is then used by to perform the required task (e.g., prediction, classification, diagnosis, etc.) on the next instance $\mathbf{u}_{n+1}$, which in turn is used to update the network and so on. In practice, the procedure might generate a new model after some number of $k$ instances are collected (see Section 4); this however, does not change the spirit of this discussion. An update procedure is evaluated by the cumulative *loss* in each step. This loss might be defined in several ways, usually depending on the particular application. Thus, for example, in classification tasks, a natural measure of this loss is the number of misclassified instances. In density estimation tasks, the intent is to measure how well the procedure predicts the next instance, and the *log-loss*, which is $\sum_n \log P_{B_n}(\mathbf{u}_n)$, is commonly used.

We start by formally describing the procedures for sequential update that define both ends of a spectrum in terms of the tradeoff between storage requirements and quality of the induced networks. We then introduce the incremental approach in Section 3.2.

### 3.1  The Naive and MAP Approaches

The naive approach to sequential update consists of storing all of the observed data, and then repeatedly invoking a batch learning procedure on $\mathbf{u}_1, \ldots, \mathbf{u}_n$ to form the estimate $B_n$. Clearly, a procedure based on this strategy uses all of the observed information to construct the next estimate and consequently should yield optimal results. However, as noted in the introduction, this approach has unreasonable space requirements in the long run. It needs to store either all of the instances that have been observed, or keep a count of the number of times each distinct instantiation to all the variables in $\mathbf{U}$ was observed. The former representation grows linearly with the number of instances collected, and will become infeasible when the network is expected to perform for long periods of time. A good example of such a network, is the "alarm" network [1] which is part of a system for monitoring of intensive care patients. In this example, the domain consists of 37 variables that can have $2^{53.95}$ distinct instantiations. Clearly, we cannot store counts for each possible instantiation observed in the data.

An alternative approach is motivated by Bayesian learning methodology. Recall that in Bayesian analysis we start with a prior probability over possible hypotheses (models and their quantifications), and compute the posterior given our observations. In principle, we can then treat this posterior as our prior for the next iteration in the sequential process. Thus, we maintain our *belief state* about the possible hypotheses after observing $\mathbf{u}_1, \ldots, \mathbf{u}_{n-1}$. Upon receiving $\mathbf{u}_n$, we compute the posterior, produce the estimate $B_{n+1}$, and store that posterior as our current belief state. This methodology has the attractive property that in the presence of some reasonable assumptions, the belief state at time $n$ is the same as the posterior after seeing $\mathbf{u}_1, \ldots, \mathbf{u}_n$ from our initial prior belief state. If we make the assumption that the structure of the network is fixed and we use *conjugate priors*, we can efficiently represent the posterior and update it after each iteration using a closed-form formula [15].

This approach, however, is infeasible when we also attempt to update the structure of the network. The BDe score is based on assumptions that allow us to compactly represent a prior using a single network and an equivalent sample size [8]. Unfortunately, the posterior cannot be compactly represented. The conjugate form for this prior essentially requires storing a *complete* network, which is equivalent to storing the counts for all possible assignments to $\mathbf{U}$.

Since we cannot realize the exact Bayesian method, we can resort to the following approximation. At each step, we find (or approximate) the *maximum a-posteriori probability* (MAP) network candidate. That is, the candidate that is considered most probable given the data seen so far. We then approximate the posterior in the next iteration by using the MAP network as the prior network with the appropriate equivalent sample size. In other words, this procedure uses the network $B_n$ as a summary of the first $n$ observations. This procedure is space efficient since we we only need to store the new instances that have been observed since we last performed the update of the MAP. An approach similar in spirit was proposed by Lam and Bacchus [10] in the context of the MDL score.

Unfortunately, by using the MAP model as the prior for the next iteration of learning, we are loosing information, and are strongly biasing the learning process toward the MAP model itself. To illustrate this phenomena, consider a scenario where U consists of two variables $X$ and $Y$. In this case, the learning procedure has to basically choose between two models: $B_1$ where $X$ is independent of $Y$, and $B_2$ where $X$ is dependent on $Y$. Now suppose that $X$ and $Y$ are indeed correlated, but after observing the first, say, 50 instances, the posterior probability of the model where $X$ is independent of $Y$ is higher. This can happen when the number of training instances is too small to determine whether observed correlations between $X$ and $Y$ are genuine correlations or artifacts of sampling noise. Since $B_1$ is the MAP model, we use it to represent our prior for the next iteration of the learning procedure. Now suppose we observe another 50 instances, and reconsider both models. At this stage, however, our prior is strongly biased toward $B_1$. Thus, even if from the 100 samples we could have determined that $X$ and $Y$ are correlated, the update strategy based on the MAP model still selects $B_1$ if the evidence for $B_2$ contained in the new instances is again weak. This phenomena becomes more pronounced as the equivalence sample size assigned to the prior grows.

### 3.2 The Incremental Learning Procedure

In this section, we propose an approach that explores the middle ground between the two extremes discussed in the previous section. Unlike the naive approach, it does not keep all possible data records (or an equivalent representation), and unlike the MAP approach, it does not rely on a single network to represent the prior information. The basic component of this procedure is a module that maintains a set $S$ of sufficient statistics records. These records allow the update procedure to select amongst a set of possible networks for the update. Before explaining the approach in detail we introduce some necessary notation. Let $\mathit{Suff}(G)$ to denote the set of sufficient statistics for $G$, that is, $\mathit{Suff}(G) = \{\hat{N}_{X_i, \mathbf{pa}(X_i)} : 1 \leq i \leq n\}$. Similarly, given a set $S$ of sufficient statistics records, let $\mathit{Nets}(S)$ to be the set of network structures that can be evaluated using the records in $S$, that is, $\mathit{Nets}(S) = \{G : \mathit{Suff}(G) \subseteq S\}$.

Suppose that we are deliberating on the choice between two structures $G$ and $G'$. As we established in Section 2, in order to use the MDL or BDe score (or variants thereof), to evaluate $G$, we need to maintain the set $\mathit{Suff}(G)$, and to evaluate $G'$, we need to maintain the set $\mathit{Suff}(G')$. Now suppose that $G$ and $G'$ differ only by one arc from $X$ to $Y$. Then there is a large overlap between $\mathit{Suff}(G)$ and $\mathit{Suff}(G')$. Namely, $\mathit{Suff}(G) \cup \mathit{Suff}(G') = \mathit{Suff}(G) \cup \{\hat{N}_{Y, \mathbf{pa}(Y)'}\}$, where $\mathbf{pa}(Y)'$ is the parent set of $Y$ in $G'$. Thus, we can easily keep track of both these structures by maintaining a slightly larger set of statistics.

To see how this generalizes to larger sets that covers a considerable subset of the search space recall that the greedy hill climbing search procedure works by comparing its current candidate $G$ to all its *neighbors*. These neighbors are the networks that are one change away (i.e., arc addition, deletion or reversal) from $G$. Extending the argument above, we see that we can evaluate the set of neighbors of $G$, by maintaining a bounded set of sufficient statistics. Note that if $S$ consists of all the sufficient statistics for $G$ and its neighbors, $\mathit{Nets}(S)$ contains additional networks, including many networks that add several arcs in distinct families in $G$. Also note that if $\mathbf{X} \subset \mathbf{Y}$, then $\hat{N}_\mathbf{X}$ can be recovered from $\hat{N}_\mathbf{Y}$. Thus, $\mathit{Nets}(S)$ also contains many networks that are simpler than $G$.

Generalizing this discussion. Our approach applies to any search procedure that can define a *search frontier*. This frontier consists of all the networks it compares in the next iteration. We use $F$ to denote this set of networks. The choice of $F$ determines which sufficient statistics are maintained in memory. That is, we set $S$ to contain all the sufficient statistics needed to evaluate the networks in $F$. After a new instance is received (or, in general, after some number of new instances are received), the procedure uses the sufficient statistics in $S$ to evaluate and select the best scoring network in the frontier $F$ (or, more generally, in $\mathit{Nets}(S)$). Once this choice is made, it invokes the search procedure to determine the next frontier, and updates $S$ accordingly. This process may start recording new information and may also remove some sufficient statistics from memory.

The main loop of the incremental procedure can be now described as follows:

```
Set G to be initial network.
Let F initial search frontier for G.
Let S = Suff(G) ∪ ⋃_{B'∈F} Suff(G').
Forever
    Read data u_n.
    Update each record in S using u_n.
    if n mod k = 0 then
        Let G = arg max_{G'∈Nets(S)} S(G' | S)
        Update the frontier F (using a search procedure)
        Set S to Suff(G) ∪ ⋃_{B'∈F} Suff(G').
    Compute optimal parameters Θ for G from S.
    Output (G, Θ).
```

This procedure focuses its resources on keeping track of just enough information to make the next decision in the search space. Every $k$ steps, the procedure performs this decision. After each such decision is made, the procedure reallocates its resources in preparation for the next iteration. This reallocation may involve removing some sufficient statistics from $S$, and adding new ones.

When we instantiate this procedure with the greedy hill climbing procedure, the frontier consists of all the neighbors of $B_n$. A beam search, on the other hand, can maintain $j$ candidates, and set the frontier to be all the neighbors of all $j$ candidates. Other search procedures might explore only some of the neighbors of $B_n$ and thus would have smaller search frontiers.

The only unresolved issue in the description of this procedure is the definition of the score $S(G \mid S)$. As we explain in the next section, we cannot directly apply the scores introduced in Section 2.

### 3.3 Scoring functions for Sequential Update

Ideally, we would like to rely on the standard scoring functions reported in the literature to score and evaluate the different structures in the procedure described above. Unfortunately, both these scores make the assumption that we are evaluating all candidates *with respect to the same dataset*. This assumption does not hold in our procedure. Since we may start collecting sufficient statistics records for different families at different times, the sufficient statistics for the family of $X$ with parent set $\mathbf{Y}$ might summarize a larger number of instances than the sufficient statistics for the parent set $\mathbf{Z}$. This effectively translates into an evaluation over different datasets.[2]

The underlying problem is a general model selection problem, where we have to compare two models $M_1$ and $M_2$ such that model $M_1$ is evaluated with respect to the training set $D_1$, while model $M_2$ is evaluated with respect to the training set $D_2$. Of course, for this problem to be meaningful, we assume that $D_1$ and $D_2$ are both sampled from the same underlying distribution. This assumption is clearly true in our case.

The MDL and the BDe scores are inappropriate for this problem in their current form. The MDL score measures the number of bits required to encode the training data if assume that the underlying distribution has the form specified by the model. However, if $D_2$ is much smaller than $D_1$, then the description of $D_2$ would usually shorter than that of $D_1$ regardless of how good the model $M_2$ is. The same problem occurs with the BDe score. This score evaluates the probability of the dataset if we assume that the underlying distribution has the form specified by the model. Again, if $D_2$ is much smaller than $D_1$, then the probability associated with it will usually be larger, since the probability of a dataset is a product of the probability of each instance given the previous ones. Since each such term is usually smaller than 1, the probability decreases for longer sequences.

We can, of course, reset all the counters every time we start gathering some new sufficient statistics record. This, in effect, restarts the learning process using the last suffix of the training sequence. This, however, would discard useful information that has been gathered about the earlier parts of the sequence. Alternatively, we can adopt the Bayesian method, and compute $P(M_1 \mid D_1)$ and $P(M_2 \mid D_2)$. Since both of these terms are degrees of belief, we can compare them and see which candidate is considered more likely given the available evidence for that candidate. Unfortunately, while we have a closed-form for $P(D_1 \mid M_1)P(M_1)$ and $P(D_2 \mid M_2)P(M_2)$, we cannot effectively evaluate the normalizing factor (which is, of course, different for $D_1$ and $D_2$). Thus, we cannot evaluate $P(M_1|D_1)$ and $P(M_2|D_2)$ in a closed-form.

We believe that the comparison of models evaluated with respect to different but related data is a fundamental problem that requires a principled solution, and we pose it as an open question. Intuitively, we want a score that assigns higher confidence to families for which we have more data.

What follows is our proposal which is based on modifying the existing MDL and BDe scores for learning Bayesian networks. This proposal satisfies a basic correctness property, and furthermore, as our experimental results show, performs well in practice.

Our proposal is best motivated in the MDL setting. This score can be casted in information theoretic terms [5, 9] as:

$$S_{MDL}(G \mid D) = N \sum_i H_D(X_i \mid \mathbf{pa}(X_i)) + \frac{\log N}{2} \sum_i \#(X_i, \mathbf{pa}(X_i)),$$

where $H_D(X_i \mid \mathbf{pa}(X_i))$ is the *empirical conditional entropy* of $X_i$ given its parents, and is equal to

$$- \sum_{x_i, \mathbf{pa}(x_i)} \frac{N(x_i, \mathbf{pa}(x_i))}{N} \log \frac{N(x_i, \mathbf{pa}(x_i))}{N(\mathbf{pa}(x_i))}.$$

As $N$ grows larger, this conditional entropy converges to the true conditional entropy in the underlying distribution:

$$H(X \mid \mathbf{pa}(X_i)) = - \sum_{x_i, \mathbf{pa}(x_i)} P(x_i, \mathbf{pa}(x_i)) \log P(x_i \mid \mathbf{pa}(x_i)).$$

This latter quantity is the smallest number of bits, for this particular underlying distribution, by which we can encode the value $X_i$ if know the value of $\mathbf{pa}(X_i)$. If we divide by $N$, we get that

$$\frac{S_{MDL}(G \mid D)}{N} \rightarrow_{N \to \infty} \sum_i H(X \mid \mathbf{pa}(X_i)) + \frac{\log N}{2N} \sum_i \#(X_i, \mathbf{pa}(X_i)).$$

In other words, the average encoding length per instance approximates the true encoding length that can be achieved with $G$. The second term in this expression embodies the *redundancy* over the optimal encoding (had we known the true distribution) incurred by the particular choice of model. For larger datasets, the redundancy in the the average encoding decreases. In effect, this redundancy term captures the amount of confidence in the learned parameters: As $N$ increases, we are more more confident in our models. At the same time, we are less confident about models that require more parameters.

This discussion suggests that the average encoding length per instance is a measure that can be compared in data sizes of different lengths. We define the *average* MDL score as

$$S^*_{MDL}(X_i, \mathbf{pa}(X_i)) = \frac{S_{MDL}(X_i, \mathbf{pa}(X_i))}{\sum_{x_i, \mathbf{pa}(x_i)} N(x_i, \mathbf{pa}(x_i))}.$$

---
[2]Note, however, that that both sufficient statistics records summarize a suffix of the sequence $\mathbf{u}_1, \ldots, \mathbf{u}_n$. Thus, one is always a subset of the other.

This score measures the average number of bits needed to represent one instance of $X_i$ given the values of $\mathbf{pa}(X_i)$. When we compare models based on different datasets, this score normalizes each evaluation to measure the average effective compression rate per instance.

The average MDL score is consistent with the original MDL score when we compare two networks given the same data. That is, when we compare two models using the same data, both scores make the same choices. To see this, note that in this case, the average MDL score assigned to the two models is their original MDL score divided by $N$. As with other scores, we also want to be ensured that in the limit the score will choose the right model. We now make this precise. We define the *inherent error* of a model $G$ with respect to a reference distribution $P^*$ to be

$$Error(G, P^*) = \min_{\Theta} D(P^* \| P_{(G,\Theta)})$$

where $D(P^* \| P_B) = \sum_{\mathbf{u}} P^*(\mathbf{u}) \log \frac{P^*(\mathbf{u})}{P_B(\mathbf{u})}$ is the cross entropy (or Kullback-Leibler divergence) between $P^*$ and $P_B$. This is a standard measure of distance between probabilistic models in statistics and information theory. We now show that given sufficient data, the average score will prefer structures that incur smaller errors.

**Lemma 3.1:** *Let $G_1$ and $G_2$ be two network structures that are evaluated with respect to datasets $D_1$ and $D_2$ of size $N_1$ and $N_2$, respectively, that are sampled i.i.d. from an underlying distribution $P^*$. If $Error(G_1, P^*) < Error(G_2, P^*)$, then as both $N_1$ and $N_2$ both go to infinity, $S^*_{MDL}(G_1 \mid D_1) < S^*_{MDL}(G_2 \mid D_2)$ with probability 1.*

We suggest a similar averaging for the Bayesian score. We define the *average* BDe score as $S^*_{BDe}(X_i, \mathbf{pa}(X_i)) = \sum_{x_i, \mathbf{pa}(x_i)} \frac{S_{BDe}(X_i, \mathbf{pa}(X_i))}{N(x_i, \mathbf{pa}(x_i))}$. This modification is motivated by the asymptotic equivalence between the Bayesian score and the MDL score. A a general result by Schwarz [14] shows that

$$S_{BDe}(G \mid D) = -S_{MDL}(G \mid D) + O(1).$$

Thus, by Lemma 3.1, the average Bayesian score is also asymptotically correct. It is not clear to us, at this stage, whether this average score has any principled probabilistic justification.

## 4  Experimental Evaluation

The experiments reported in this section were designed with two objectives in mind. First, to show the performance of the methods we described in the previous section. Second, to evaluate the tradeoff between space used by the learning procedure and the quality of the learned networks.

We compared the performance of the three procedures we described above. All three procedures take as a parameter the number $k$ of instances after which the network structure is reconsidered. The procedures are:

- **Naive**– the naive approach described in Section 3.1. This procedure stores all the instances seen so far, and invokes a greedy hill climbing search procedure every $k$ instances to select a network.

- **MAP**– the MAP procedure described in Section 3.1. In this procedure the best network so far is used as the prior for the update in the next iteration (consisting of $k$ instances). The procedure also use greedy hill climbing to select a network.

- **Incremental**– the procedure introduced in Section 3.2. Every $k$ instances, the procedure uses a on greedy hill climbing search to select a network from $Nets(S)$, the set of networks that can be evaluated using the currently stored statistics. This procedure uses the neighbor frontier set we described above.

The datasets used in the experiments were generated from two networks: the **alarm** network of [1] and the **insurance** network of [13]. The alarm networks contains 37 variables, and the insurance network contains 26 variables. From each network we sampled 5 training sets, each consisting of 10,000 instances. The results reported in Figures 1 and 2 are averages over the results of running the algorithms on all 5 datasets.

For all three procedures described above we update the parameters quantifying the current network candidate after each instance is received. We set the prior probability to be a weak uniform prior (equivalent sample size of 5). We run tests for the following values of $k$ 100, 400 and 800. As expected, the performance of **Naive** is mostly independent of $k$. Thus, we report the result for $k = 100$ only. We run **Naive** with both the MDL and BDe scores. Similarly, we run **Incremental** with both the modified scores introduced in Section 3.3.

The evaluation of the quality of the induced networks is based on the *log-loss*, that is $\sum_n \log P_{B_n}(\mathbf{u}_n)$. This is a standard measure of performance in density estimation. Since we generated the training sets from existing networks, we can measure how close the learned models are to the generating distribution. In particular, we measured the *normalized loss* $\sum_n (\log P^*(\mathbf{u}_n) - \log P_{B_n}(\mathbf{u}_n)) = \sum_n \log \frac{P^*(\mathbf{u}_n)}{P_{B_n}(\mathbf{u}_n)}$, where $P^*()$ is the probability of the generating distribution. This measure has several attractive properties. First, it relates the loss incurred by the learning procedure to the optimal that can be reached. Thus, it measures the additional penalty for using an approximation instead of the true model. Second, the normalized loss is related to the cross-entropy measure of distance between probabilistic models. It is easy to see that cross entropy can be rewritten as $D(P^* \| P_{B_n}) = E[\log \frac{P^*(\mathbf{u})}{P_{B_n}(\mathbf{u})}]$, where the average is based on $P^*$. Thus, the average normalized loss is an estimate of the cross entropy.

Figure 1 summarizes the performance of each method in terms of the average normalized loss. As expected, **Naive** does better than all the others, since it is able to use all the information observed in the data in a cumulative way. The figures do not show significant difference between the MDL and BDe scores for this procedure.

As we anticipated (Section 3.1), we observe **MAP**'s ten-dency to lock on its current model once the prior received sufficient weight. This can be seen very clearly from the results of Figure 1. The qualitative behavior of **MAP** approach shows improvement, until a stage where the equivalent sample size of the prior is larger than about $5 * k$, then the procedure locks on a particular structure and the performance levels off. In the case of $k = 100$, the performance actually degraded after this stage. (Unfortunately the performance of this case falls off the range of the graphs.) Of course, we might deal with this problem by setting the prior strength to be smaller than the number of examples seen. This, however, runs the risk of learning too simple networks. (Recall that we adopt complex networks only if the data supports it in a sufficiently large number of samples.) We are currently in the process of experimenting with different strategies of setting the prior's equivalent sample size.

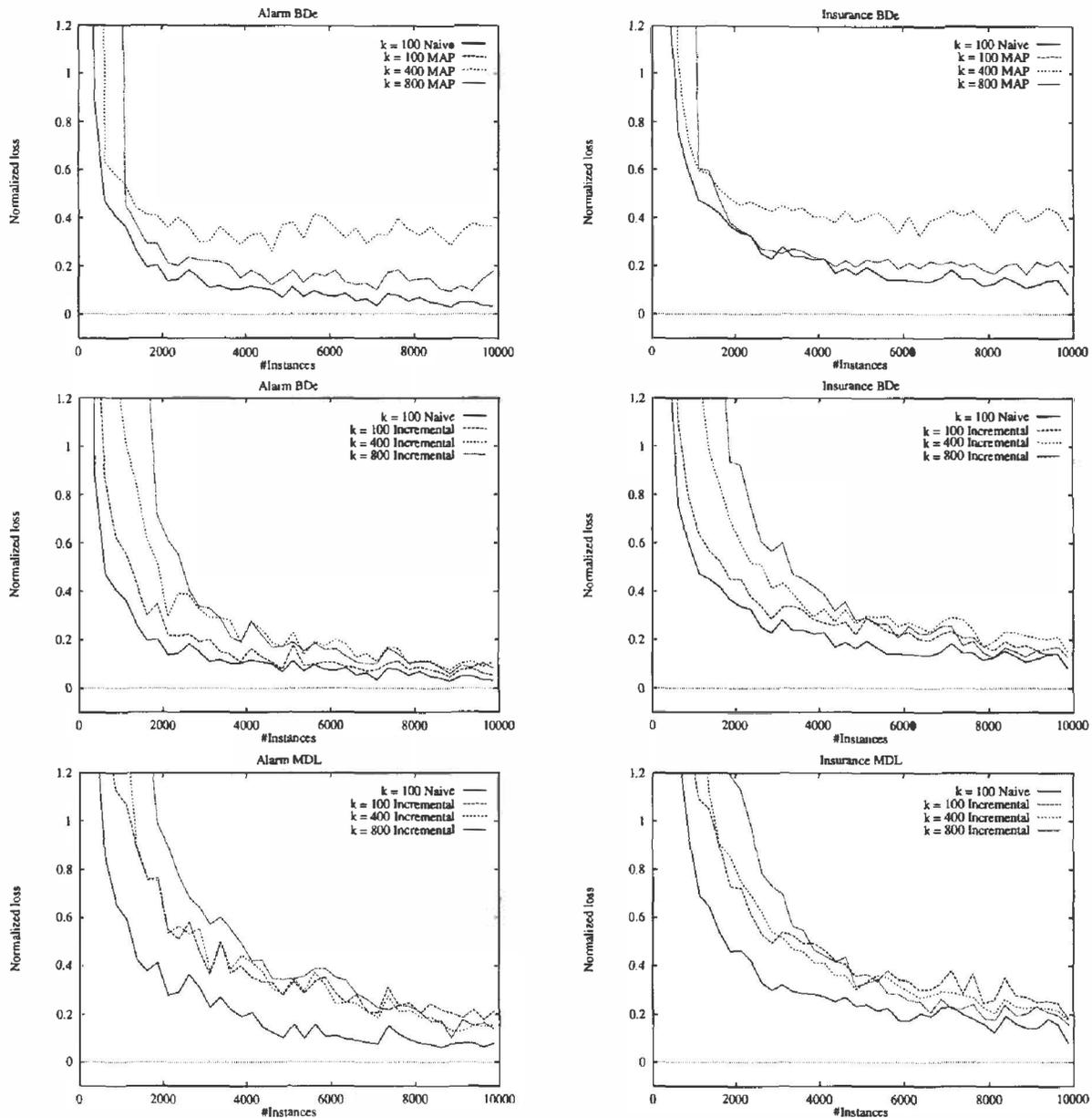

Figure 1: Performance results of several methods for sequential update. The left column reports results for "alarm" network and the right column reports results for the "insurance" network. Results in the the top row use the MAP procedure with the BDe score, the middle row use the incremental procedure using the BDe score, and the lower row use the incremental procedure with the MDL. All the graphs also display the results of the naive method with the corresponding score as a point of reference. The horizontal axis measures the number of instances seen. The vertical axis measures the average normalized loss, where each data point is the average of a window of 250 instances over 5 datasets.

As described in Section 3.2, we expect **Incremental** to converge to the right distribution, albeit more slowly than **Naive**. This is indeed the qualitative behavior we can see in Figure 1. These curves show that the scores are consistent and that the incremental procedure is sound, in the sense that as more data is observed, the procedure outputs networks that are closer in performance to the golden model and the best possible sequential learning exhibit by the naive procedure. Also as expected, runs with larger $k$ needed more time to achieve a good level of performance. This is due to the fact that the procedure posts new sufficient statistics records only every $k$ instances, and thus, it needs several $k$ instances before it can start maintaining sufficiently complex statistics. However, in the long run, the procedures with larger value of $k$ seem more robust.

Figure 2 summarizes the space usage for these procedures. The estimates are optimistic in that we count only the necessary data structures and not any auxiliary ones. For **Naive** and **MAP**, we measure the space needed to store the different instances. For **Incremental**, we measure the space needed to store all of the active sufficient statistics records. Again as expected, **Naive** requires more and more memory since it needs to store all the previously seen instances. On the other hand, **MAP** requires a constant amount of memory, since it stores at most $k$ instances at any time. Finally, **Incremental** allocated more memory as it learned a more complex structure. However, once the structure estimate stabilized, the memory usage of this procedure also stabilized and remained roughly constant.

We remark that the memory usage in **Incremental** can be minimized further when we have prior knowledge that limits the space of possible networks (e.g., ordering constraints). Additionally, we can use heuristic estimates to estimate which arc additions are the most promising (using, for example, pairwise mutual information), and only maintain the corresponding records. We suspect that such a scheme can reduce the space requirements by a large factor at a small penalty in performance. We are currently in the process of experimenting with these extensions.

## 5 Missing Data

In the above discussion, we have made the assumption that the data is complete in the sense that each data instance $u_n$ contains values for each variable in U. Unfortunately, in many real life applications we are forced to deal with incomplete data. The source of these incompleteness may come from noisy measurements, or from domains in which some attributes are not directly observed.

One source of difficulty in learning from incomplete data is computational in nature. We can no longer decompose the probability of the data. This means that the score (either MDL or Bayesian) cannot be written as the sum of local terms measuring how well we model the probability of each variable given its parents as in Equation 1. Moreover, in order to evaluate the optimal choice of parameters for a given candidate network structure, we must perform a nonlinear optimization using either *Expectation Maximization* EM [11] or gradient descent [13]. In this paper we focus on the EM procedure. The standard use of EM is for batch learning. In addition, it is restricted to induce the parameters under the assumption of a fixed structure. In order to adapt EM to the sequential update problem we need to relax both these restrictions. Fortunately, two recent methods deal with each of these restrictions in turn, and as we now describe, a combination of both methods leads to an elegant learning procedure for sequential update.

For reasons of space, we keep the discussion at a high level, and refer the interested reader to the relevant papers cited below. The standard EM procedure for learning parameters iteratively and monotonically improves its current choice of parameters for a fixed structure $G$ using the following two steps. In the first step, the current parameters $\Theta$ are used for computing the *expected* value of all the relevant sufficient statistics records, $\textit{Suff}(G)$. This computation evaluates $E[N(\mathbf{x}) \mid D, P_B] = \sum_C N_{D^+}(\mathbf{x}) P_B(D^+ \mid D)$, where $B = (G, \Theta)$ and we sum over all the possible *completions* $D^+$ of $D$ (that is, assignments to all unobserved values in $D$). In the second step, $\Theta$ is replaced by the parameters $\Theta'$ estimated from those expected statistics. This second step is essentially equivalent to learning parameters from complete data. The theoretical justification for this procedure shows that by proceeding in this manner with each iteration we increase the probability that the observed data was sampled from the distribution represented by the structure and the induced parameters.

As described the procedure is a batch learning method in that we must retain all of the training set. Neal and Hinton [12] essentially show that we also improve this probability if we use an incremental update of the sufficient statistics. In their approach, new incoming data cases are used to continuously recompute the sufficient statistics. The intuitive justification behind this approach is that any two sufficiently long sequence of of i.i.d. samples are similar. Thus, instead of storing the training data, this procedure relays on new data. The second enhancement they describe justifies updates to the parameters after every instance is seen. This results in the following procedure for learning parameters:

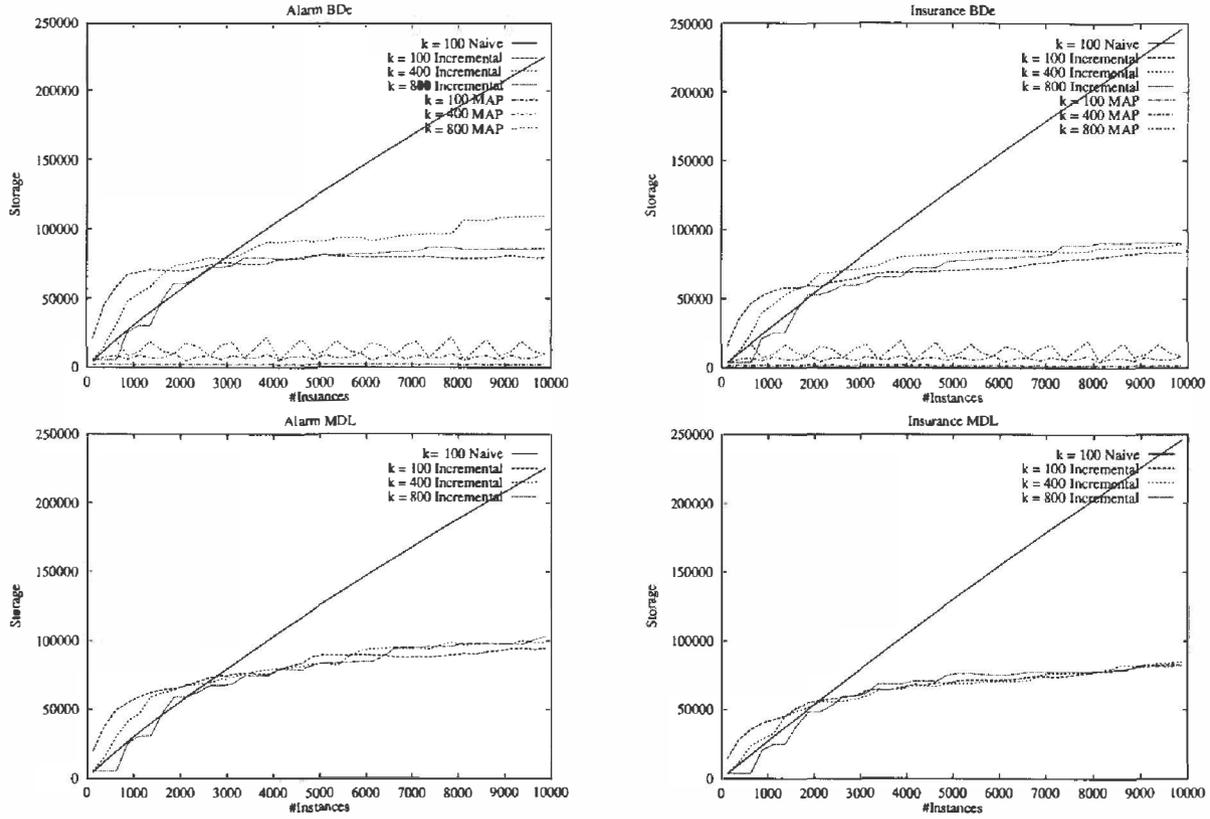

Figure 2: Space requirements of several methods for sequential update. The upper row reports methods using the BDe score and the lower methods using the MDL score. The left column reports results for the "alarm" network and the right columns reports results for the "insurance" networks. The horizontal axis measures the number of instances seen. The vertical axis measures the average space needed to store the data retained by the procedure.

Set $B = (G, \Theta)$ to be initial network.
For all $\hat{N}_{\mathbf{X}} \in \mathit{Suff}(G)$
    $N(\mathbf{x}) = N_0 \cdot P_B(\mathbf{x})$
Forever
    Read data instance y.
    For all $\hat{N}_{\mathbf{X}} \in \mathit{Suff}(B)$
        $N(\mathbf{x}) \leftarrow N(\mathbf{x}) * \alpha + P_B(\mathbf{x}|\mathbf{y})$
    Update parameters in $\Theta$ from the new sufficient statistics.
    Output $B$

In this procedure, $N_0$ designates the confidence in the initial network, and $\alpha$ is a *decay* parameter, which has a value less than 1. Usually we set $\alpha$ to be quite close to 1, e.g., 0.99. Using this decay parameter we gradually decrease the contribution of old samples (which where "completed" by old parameters). This procedure is not guaranteed to make positive progress in each step, however, on the average, it does make progress and for a sufficiently long sequence, the procedure will converge.

The second restriction of the standard EM is that it deals only with learning parameters in a fixed structure. Friedman [4] shows that if use the expected sufficient statistics to evaluate alternative structures using the MDL score and choose structures that are assigned a higher score than the current model, then we are bound to improve the marginal score of network with respect to the observed data. Simi-lar results apply to the Bayesian score as well. It follows then that we can use the expected sufficient statistics in our search procedure to evaluate new models.

There is no intrinsic difficulty in casting the results of [4] in the incremental framework of [12]. Combining the two techniques we get a simple modification of our approach that deals with incomplete data:

Set $B$ to be initial network.
Let $F$ initial search frontier for $B$.
Let $S = \mathit{Suff}(B) \cup \bigcup_{B' \in F} \mathit{Suff}(B')$.
For all $\hat{N}_{\mathbf{X}} \in S$
    $N(\mathbf{x}) \leftarrow N_0 \cdot P_B(\mathbf{x})$
Forever
    Read data instance y.
    For all $\hat{N}_{\mathbf{X}} \in S$
        $N(\mathbf{x}) \leftarrow N(\mathbf{x}) * \alpha + P_B(\mathbf{x}|\mathbf{y})$
    if $n \bmod k = 0$ then
        Let $B = \arg\max_{B' \in \mathit{Nets}(S)} S(B' \mid S)$
        Update the frontier $F$
        Update $S$ to $\mathit{Suff}(B) \cup \bigcup_{B' \in F} \mathit{Suff}(B')$.
    Compute optimal parameters for $B$ from $S$.
    Output $B$.

It is well known that EM methods may reach a sub-optimal local minima. Standard techniques to avoid these minima such as, for example, running EM several times with different random starting points apply here as well. Of course, we can run parallel executions of this procedure. We are currently in the process of evaluating the effectiveness of this procedure.

## 6 Discussion

Previous work on the sequential update of Bayesian networks have been mostly restricted to updating the parameters assuming a fixed structure [15]. The two notable exceptions are the approaches by Buntine [2] and by Lam and Bacchus [10]. Buntine's method assumes that a total order on the variables is given, and it maintains sufficient statistics for the possible parents of each node using lattice structures. He imposes restrictions on the size of the lattices in order to bound the amount of information that is maintained. Unfortunately, Buntine does not provide any reports on experimental evaluations of his approach which makes a rigorous comparison difficult. Lam and Bacchus propose a different approach based on a modification of the MDL score. In essence, they use the current network in each iteration as a summary of previously seen data. This is similar, in spirit, to the MAP approach described in Section 3.1. They performed experiments with the alarm network, yet their evaluation criteria is not objective in the sense of a log-loss scoring of the resulting networks, and once more, rigorous comparisons are difficult.

The incremental approach introduced in this paper opens new degrees of flexibility in various dimensions. First, by using different search strategies, different search frontiers are achieved, which in turn changes the amount of information maintained. Additionally, we are actively considering different heuristics for pruning the number of sufficient statistics stored in memory. These heuristics would rule out statistics that are unlikely to lead to an arc addition.

As mentioned in the introduction, the approach we describe is useful in applications that involve large amounts of data. Such applications include data mining problems and monitoring problems. We also note that by keeping a set of likely candidates readily available, this approach can also be useful to cases where there are real-time constraints on the model selection process, and decisions have to be made in a real-time fashion.

There are two topics that we are currently exploring. The first is an empirical evaluation of these three methods in situations where the underlying distribution drifts in time. Our goal is to characterize the different parameters that affect the efficiency of the procedure in "tracking" these changes. The second topic involves learning in the presence of incomplete data. We established the foundations of a solution with the procedure described in Section 5. We are in the process of conducting experiments that will allow us to evaluate the approach and propose further refinements if needed.

Finally, even though our results confirm the consistency and effectiveness of the scores introduced in Section 3.3, we would like a principled derivation based on a strictly probabilistic interpretation. As mentioned in that section, we pose this as an open problem.

### Acknowledgments

Parts of this work were done while Nir Friedman was at SRI International. Nir Friedman's work at Berkeley was funded by ARO through MURI DAAH04-96-1-0341 and by NSF through FD96-34215.